\pdfoutput=1

\documentclass{article}

\usepackage{arxiv}

\usepackage[utf8]{inputenc}
\usepackage[T2A,T1]{fontenc}
\usepackage{hyperref}
\usepackage{url}
\usepackage{booktabs}
\usepackage{amsfonts}
\usepackage{microtype}
\usepackage{graphicx}
\usepackage{natbib}
\usepackage{enumitem}
\usepackage{amsmath}
\usepackage[noblocks]{authblk}
\newcommand{\headeright}{}
\newcommand{\undertitle}{}

\title{Subword-Based Comparative Linguistics across 242 Languages\\Using Wikipedia Glottosets}

\author[1,2,3]{\textbf{Iaroslav Chelombitko}}
\author[3]{\textbf{Mika H\"am\"al\"ainen}}
\author[2,4]{\textbf{Aleksey Komissarov}}

\affil[1]{DataSpike} 
\affil[2]{Neapolis University Pafos, \textbf{Paphos, Cyprus}}
\affil[3]{Metropolia University of Applied Sciences, \textbf{Helsinki, Finland}}
\affil[4]{aglabx, \textbf{Paphos, Cyprus}}
\affil[ ]{\texttt{i.chelombitko@nup.ac.cy}}
\date{}

\renewcommand{\shorttitle}{Subword-Based Comparative Linguistics across 242 Languages}

\hypersetup{
pdftitle={Subword-Based Comparative Linguistics across 242 Languages Using Wikipedia Glottosets},
pdfsubject={cs.CL},
pdfauthor={Iaroslav Chelombitko, Mika Hamalainen, Aleksey Komissarov},
pdfkeywords={Comparative Linguistics, BPE, Tokenization, Multilingual NLP, Language Phylogeny},
}

\begin{document}
\maketitle

\begin{abstract}
We present a large-scale comparative study of 242 Latin and Cyrillic-script languages using subword-based methodologies. By constructing 'glottosets' from Wikipedia lexicons, we introduce a framework for simultaneous cross-linguistic comparison via Byte-Pair Encoding (BPE). Our approach utilizes rank-based subword vectors to analyze vocabulary overlap, lexical divergence, and language similarity at scale.
Evaluations demonstrate that BPE segmentation aligns with morpheme boundaries 95\% better than random baseline across 15 languages (F1 = 0.34 vs 0.15). BPE vocabulary similarity correlates significantly with genetic language relatedness (Mantel r = 0.329, p < 0.001), with Romance languages forming the tightest cluster (mean distance 0.51) and cross-family pairs showing clear separation (0.82). Analysis of 26,939 cross-linguistic homographs reveals that 48.7\% receive different segmentations across related languages, with variation correlating to phylogenetic distance.
Our results provide quantitative macro-linguistic insights into lexical patterns across typologically diverse languages within a unified analytical framework.

\end{abstract}

\keywords{Multilingual NLP \and Comparative Linguistics \and BPE Tokenization \and Cross-linguistic Analysis \and Lexical Similarity \and Script-based Typology}

\section{Introduction}

Traditional comparative linguistics, while providing deep historical and typological insights (see \citealt{lehmann2013historical, beekes2011comparative, Nichols1992LinguisticDI, partanen2021processing, saily2021plenipotentiary}), often lacks the scalability to handle the current volume of digital text (\citealt{arnett2024languagemodelsperformworse, akindotuni2025}). However, recent NLP advances facilitate massive, data-driven studies (see \citealt{dang2024ayaexpansecombiningresearch, bender2011languageindependence, imanigooghari-etal-2023-glot500, ostling2016continuous}), revealing universal tendencies in phonetics (\citealt{blum2024consonant}), lexical semantics (\citealt{tjuka2024universal}), and sound symbolism (\citealt{cwiek2022bouba, cathcart2024sound}) that were previously inaccessible through manual, small-scale methods.

Despite this progress, large-scale studies frequently neglect endangered languages (\citealt{hamalainen2021endangered}) and systematic performance disparities in multilingual models (see \citealt{shani2026rootsperformancedisparitymultilingual, chelombitko2024qtokcomprehensiveframeworkevaluating, dunn2011evolved}). We address this gap by examining 242 languages through a novel script-level lens (Latin vs.\ Cyrillic), moving beyond family-specific comparisons (\citealt{meillet1967comparative, beekes2011comparative}) to reveal overarching patterns visible only when languages share a script-based writing system (\citealt{daniels2007world, rust-etal-2021-good}).

In particular, we adopt a subword-based strategy (\citealt{sennrich2016neural}) using Byte-Pair Encoding (BPE) (\citealt{10.5555/177910.177914}) tokenizers, which we train both on individual languages and on large aggregated corpora for each script. Aggregating data for all Latin and Cyrillic-script languages into respective training sets effectively unifies each script community into a single model, leveraging the shared-parameter paradigm common in massively multilingual pretraining (see \citealt{johnson2016google, conneau-etal-2020-unsupervised}).

\begin{figure}[ht]
  \centering
  \includegraphics[width=\textwidth]{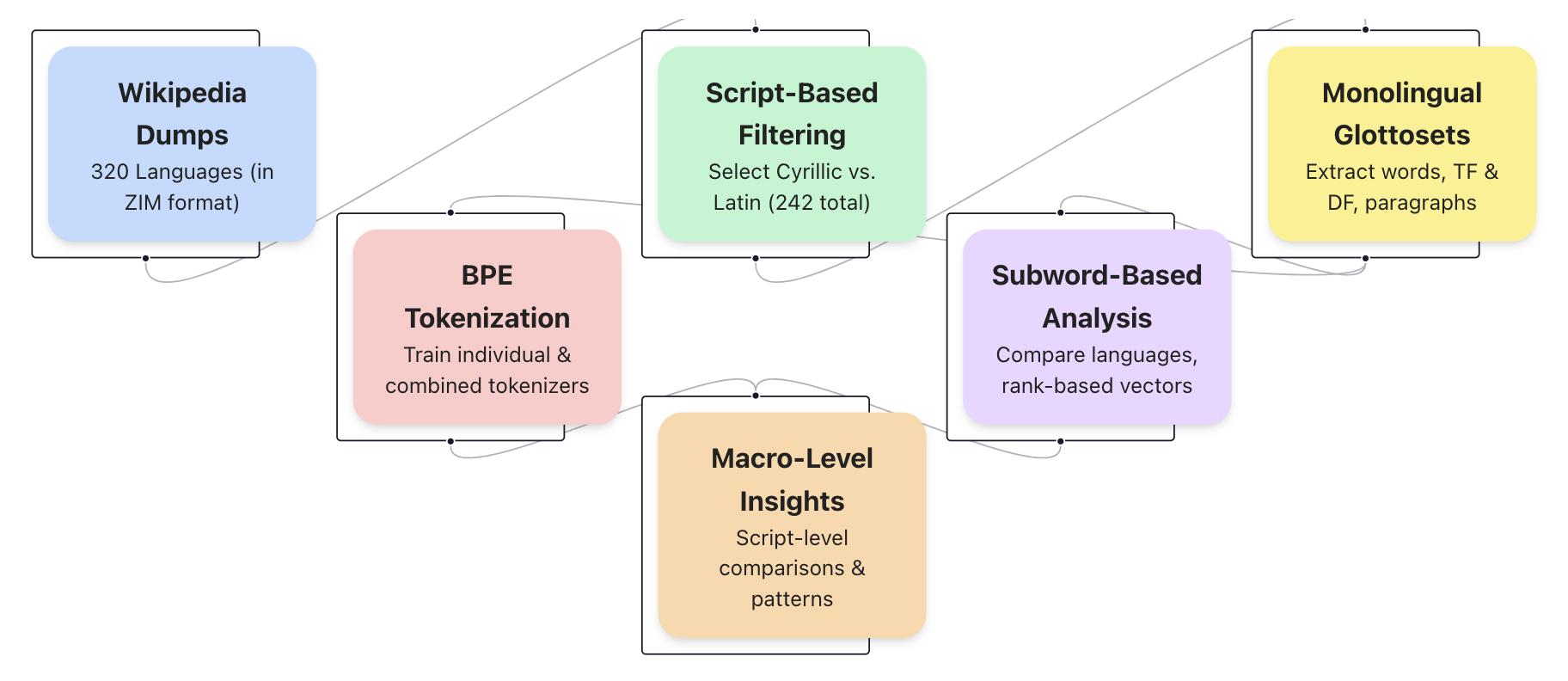}
  \caption{Pipeline architecture for subword-based comparative linguistics across 242 languages using Wikipedia glottosets. The workflow illustrates the transformation of Wikipedia dumps (320 languages) through sequential stages: script-based filtering yielding 37 Cyrillic and 205 Latin script languages, monolingual glottoset construction with TF/DF metrics, BPE tokenization (both individual and combined training with 4096 tokens vocabulary), and vector-based subword analysis. Each colored node represents a distinct processing phase, culminating in macro-level insights for script-level comparative linguistics. This modular approach enables scalable analysis of morphological patterns across multiple languages simultaneously.}
  \label{fig:workflow}
\end{figure}

Although subword units are not perfect morphemes (see \citealt{sennrich2016neural, bostrom2020bytepairencodingsuboptimal}), they serve as robust automated tools for macro-linguistic research (see \citealt{khurana2024crosslingual, pham-etal-2024-unibridge, futrell2015dependency}) that scale more efficiently than manual expert alignment (see \citealt{campbell2020, rama2018automatic, ciobanu2014cognates}).

Leveraging the framework (Figure~\ref{fig:workflow}), this work makes two key contributions:
\begin{enumerate}
\item \textbf{Script-Level Aggregation:} We demonstrate how combining languages by script (Latin vs.\ Cyrillic) enables macro-level comparative insights difficult to achieve through traditional language-by-language lenses.
\item \textbf{Practical Subword Methodology:} We introduce a tractable, automated framework that reduces reliance on manual annotation by providing data-driven lexical segments for robust cross-linguistic analysis.
\end{enumerate}

While traditional comparative linguistics focuses on genetic relationships and historical reconstruction, and typology examines structural similarities regardless of ancestry, our subword-based approach bridges both: we demonstrate that BPE tokenization captures phylogenetic signal (Section~\ref{sec:e3}) while also revealing typological patterns across unrelated languages sharing the same script.

\section{Related Work}

Comparative linguistics in NLP has evolved from general data-driven frameworks (see \citealt{bender2011languageindependence, ostling2016continuous, imanigooghari-etal-2023-glot500}) to specialized tasks like automated cognate detection (\citealt{ciobanu2014cognates}). While manual expert annotation remains the high-precision gold standard (\citealt{Nichols1992LinguisticDI}), \citet{rama2018automatic} demonstrated that automated methods can reconstruct language phylogenies with accuracy closely approximating expert-curated data (\citealt{Jger2018GlobalscalePL}). These approaches highlight the potential for scaling linguistic analysis across diverse language families where manual examination is impractical.

To overcome the scarcity of parallel data in low-resource settings, researchers have utilized neural machine translation (NMT) to infer cognate relationships (\citealt{h2019finding}). This direction has proven particularly effective for endangered Uralic languages, leveraging cross-lingual relations (see \citealt{partanen2021processing, chelombitko-komissarov-2024-specialized, chelombitko-etal-2025-samponlp}) and data augmentation through synthetic cognates generated by statistical machine translation (SMT) (\citealt{poncelas2019combiningsmtnmtbacktranslated}). Such hybrid strategies address resource limitations while enriching existing linguistic databases for under-represented script communities.

More recently, architectures inspired by computational biology have addressed previous computational bottlenecks. \citet{akavarapu2024automated} introduced transformer-based models that utilize multiple sequence alignments and link prediction components, mirroring techniques used in genomic research. By treating linguistic evolution through an end-to-end lens, these models significantly reduce computation time while maintaining high accuracy (\citealt{doi:10.1073/pnas.1204678110}), providing a scalable alternative to traditional alignment-based methods.

Automated language typology remains another central research vein (\citealt{ponti-etal-2019-modeling}), with studies using information-theoretic measures to quantify morphological synthesis and fusion (see \citealt{rathi-etal-2021-information, oncevay-etal-2022-quantifying}). Crucially, \citet{gutierrez-vasques-etal-2023-languages} found that BPE compression ratios directly correlate with morphological complexity. This aligns with broader efforts to leverage cross-lingual representations that capture typological relationships even without parallel data (\citealt{yu2021language}), although the distribution of this linguistic information varies across model layers based on pretraining strategy (\citealt{choenni2022investigating}).

\section{Methodology}

\subsection{Monolingual Glottosets Construction}

We downloaded Wikipedia dumps in ZIM format for 320 languages available in the ZIM archive\footnote{\url{https://dumps.wikimedia.org/kiwix/zim/wikipedia/}}.
After that, we selected either the 2024 or 2023 version (preferring the 2024 version if available), otherwise, we used the 2023 version. We included all topics by selecting the ``all'' files option. To clean the data from unnecessary HTML elements, we used a custom script available in our repository. In brief, this script resolves redirects and extracts only paragraph text while removing as many auxiliary HTML tags as possible, leaving only clean paragraph texts.

\begin{figure}[ht]
  \centering
  \includegraphics[width=\textwidth]{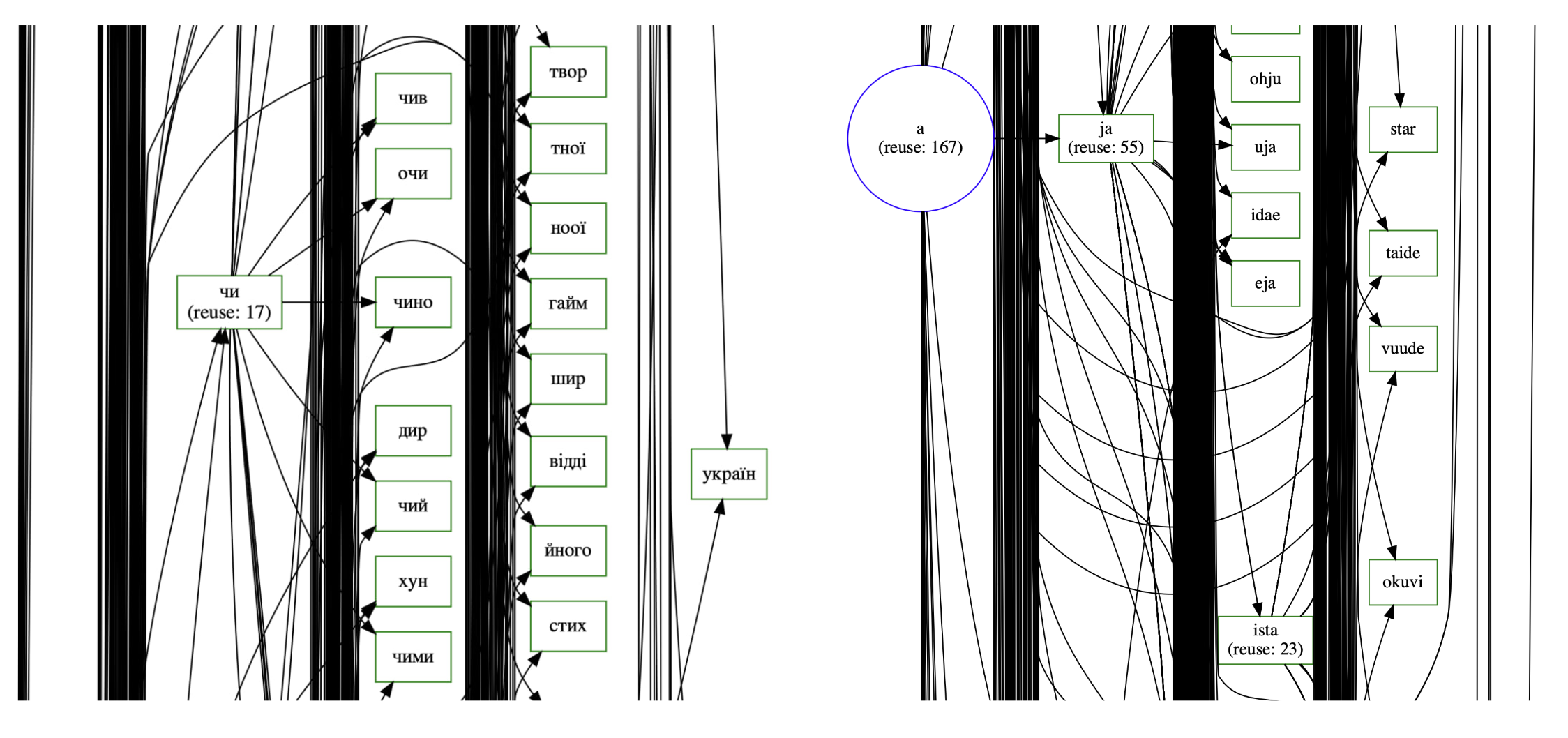}
  \caption{Visualization of interactive BPE merge graphs for Ukrainian (left) and Finnish (right) subword tokenization patterns. The diagrams show directed merge sequences with reuse counts indicated in parentheses. Vertical black bars represent merge steps, while edges show the progression of subword unit formation. The contrasting patterns reflect language-specific morphological characteristics: Ukrainian showing consistent Cyrillic character combinations, while Finnish exhibits agglutinative patterns. Interactive web application available on our repository.}
  \label{fig:bpe_merge}
\end{figure}

The extracted paragraphs were filtered based on script consistency. For Cyrillic, we retained only Cyrillic paragraphs, and for Latin, only those written in the Latin script. Unfortunately, there is currently no reliable tool for accurately detecting all 242 languages, so at this stage, we filtered only by script (Table~\ref{tab:data_stats}). Additionally, an effective filtering step was to remove all paragraphs containing fewer than 10 words. This significantly reduced noise in the dataset, including accidental words from other languages.

\begin{table}[ht]
    \caption{Statistics for top 30 languages by token count from Wikipedia dumps. Columns show language names and codes, number of raw and cleaned paragraphs, word count in millions, script type (Latin/Cyrillic), total tokens, and cross-script contamination (Cyrillic tokens in Latin-script languages and vice versa).}
    \label{tab:data_stats}
    \centering
    \small
    \renewcommand{\arraystretch}{1.1}
    \begin{tabular}{l c c c c c c c c}
        \toprule
        \textbf{Lang} & \textbf{Code} & \textbf{Raw pars} & \textbf{Clean pars} & \textbf{Words (M)} & \textbf{Script} & \textbf{Tokens} & \textbf{Cyr.} & \textbf{Lat.} \\
        \midrule
        English & en & 54125720 & 42149266 & 2661.93 & Latin & 2661930466 & 5062 & -- \\
        German & de & 22354272 & 17953642 & 1026.43 & Latin & 1026430544 & 2648 & -- \\
        French & fr & 20464072 & 16742203 & 940.75 & Latin & 940746868 & 1621 & -- \\
        Spanish & es & 14107859 & 12170134 & 763.36 & Latin & 763357750 & 1134 & -- \\
        Italian & it & 11917473 & 9806341 & 571.76 & Latin & 571759224 & 1150 & -- \\
        Russian & ru & 15290408 & 12295838 & 552.42 & Cyrillic & 552416889 & -- & 429004 \\
        Cebuano & ceb & 16841937 & 10069736 & 447.85 & Latin & 447850263 & 92 & -- \\
        Portuguese & pt & 6845897 & 5642228 & 332.96 & Latin & 332959017 & 413 & -- \\
        Dutch & nl & 7764590 & 6429095 & 316.21 & Latin & 316210100 & 415 & -- \\
        Polish & pl & 15961541 & 5616831 & 253.85 & Latin & 253846410 & 2044 & -- \\
        Catalan & ca & 4273906 & 3749516 & 238.81 & Latin & 238813199 & 710 & -- \\
        Ukrainian & uk & 7864058 & 5582991 & 219.83 & Cyrillic & 219827684 & -- & 200945 \\
        Vietnamese & vi & 4707518 & 3129629 & 201.38 & Latin & 201383401 & 623 & -- \\
        Swedish & sv & 7437772 & 5088656 & 200.89 & Latin & 200889695 & 312 & -- \\
        Serbo-Croatian & sh & 2248318 & 1682017 & 193.56 & Latin & 193561280 & 19539 & -- \\
        Czech & cs & 3386803 & 2865556 & 150.39 & Latin & 150393063 & 998 & -- \\
        Hungarian & hu & 3932082 & 2603906 & 129.86 & Latin & 129864098 & 518 & -- \\
        Indonesian & id & 2984643 & 2120278 & 106.98 & Latin & 106980811 & 622 & -- \\
        Norwegian Bokm\aa l & nb & 2906506 & 2239883 & 105.51 & Latin & 105507751 & 175 & -- \\
        Finnish & fi & 3141975 & 2295657 & 93.5 & Latin & 93502432 & 650 & -- \\
        Serbian & sr & 3210579 & 1981413 & 91.81 & Cyrillic & 91807037 & -- & 193621 \\
        Turkish & tr & 2528498 & 1757628 & 85.16 & Latin & 85156035 & 570 & -- \\
        Romanian & ro & 2360641 & 1539046 & 84.62 & Latin & 84617897 & 454 & -- \\
        Bulgarian & bg & 1599763 & 1386917 & 69.12 & Cyrillic & 69123830 & -- & 36265 \\
        Waray & war & 2571970 & 1448198 & 69.04 & Latin & 69038926 & 0 & -- \\
        Danish & da & 1551416 & 1216805 & 61.11 & Latin & 61107267 & 126 & -- \\
        Galician & gl & 1188136 & 1049404 & 60.95 & Latin & 60947590 & 181 & -- \\
        Malay & ms & 1499560 & 1034055 & 56.37 & Latin & 56370979 & 106 & -- \\
        Asturian & ast & 1134182 & 912479 & 56.25 & Latin & 56254683 & 84 & -- \\
        Esperanto & eo & 1669507 & 1188077 & 54.51 & Latin & 54512106 & 441 & -- \\
        \bottomrule
    \end{tabular}
\end{table}

In the following analysis we included only languages with Cyrillic (37 languages) and Latin (205 languages) scripts. The decision to restrict the scope to Latin and Cyrillic script languages was motivated by three primary factors: (1) these scripts share a common alphabet origin, enabling direct character-level comparison; (2) they represent the largest script families in Wikipedia coverage; and (3) they include both closely related language families (e.g., Romance, Slavic) and typologically diverse languages (e.g., Finnish, Turkish, Vietnamese), providing a rich testbed for subword-based comparative analysis. Additionally, as an unexpected result, we were able to measure the extent to which Wikipedia articles are contaminated with Latin and Cyrillic scripts for languages that use other scripts (80 languages). The results are presented in Table~\ref{tab:additional_languages}.

\begin{table}[ht]
    \caption{Script contamination analysis for languages using non-Latin/non-Cyrillic writing systems. Data shows language names, codes, raw paragraph counts, native script types, and the number of detected Latin and Cyrillic tokens representing potential script contamination.}
    \label{tab:additional_languages}
    \centering
    \small
    \renewcommand{\arraystretch}{1.1}
    \begin{tabular}{l c c c c c}
        \toprule
        \textbf{Lang} & \textbf{Code} & \textbf{Raw pars} & \textbf{Script} & \textbf{Cyrillic} & \textbf{Latin} \\
        \midrule
        Min Nan Chinese & nan & 830227 & Han & 13 & 541415 \\
        Japanese & ja & 12737874 & Japanese & 915 & 298430 \\
        Chinese & zh & 7206035 & Han & 554 & 117304 \\
        Greek & el & 1733250 & Greek & 363 & 96009 \\
        Hebrew & he & 3368324 & Hebrew & 479 & 90907 \\
        Korean & ko & 3333575 & Hangul & 635 & 81612 \\
        Arabic & ar & 4624040 & Arabic & 246 & 58738 \\
        Persian & fa & 3132901 & Arabic & 165 & 51729 \\
        Balinese & ban & 87175 & Balinese & 4 & 48922 \\
        Thai & th & 944565 & Thai & 305 & 43559 \\
        Armenian & hy & 1647293 & Armenian & 2579 & 41259 \\
        Egyptian Arabic & arz & 5166127 & Arabic & 23 & 39349 \\
        Buginese & bug & 36776 & Lontara & 0 & 19834 \\
        Min Dong Chinese & cdo & 28589 & Han & 0 & 19197 \\
        Georgian & ka & 644781 & Georgian & 387 & 15234 \\
        Bengali & bn & 854059 & Bengali & 64 & 13786 \\
        Hindi & hi & 703141 & Devanagari & 80 & 13369 \\
        Tamil & ta & 820809 & Tamil & 28 & 11153 \\
        Hakka Chinese & hak & 14204 & Han & 2 & 10368 \\
        Malayalam & ml & 464433 & Malayalam & 31 & 9848 \\
        Sinhalese & si & 166475 & Sinhala & 3 & 9552 \\
        Burmese & my & 451126 & Burmese & 8 & 9467 \\
        Urdu & ur & 660706 & Arabic & 62 & 9172 \\
        Cantonese & yue & 373575 & Han & 53 & 8418 \\
        South Azerbaijani & azb & 605161 & Arabic & 18 & 7459 \\
        Goan Konkani & gom & 39569 & Devanagari & 0 & 6955 \\
        Marathi & mr & 349639 & Devanagari & 10 & 6842 \\
        Cambodian & km & 99641 & Khmer & 13 & 6324 \\
        Telugu & te & 832028 & Telugu & 3 & 6170 \\
        Tachelhit & shi & 7811 & Tifinagh & 0 & 5799 \\
        \bottomrule
    \end{tabular}
\end{table}

We define monolingual glottosets as language-specific collections of lexical units derived from monolingual texts. Each glottoset reflects a particular language's vocabulary in one script, in our case Latin or Cyrillic in lowercase. Unfortunately, while we initially aimed to avoid normalizing words to lowercase, we found that doing so significantly improves subsequent tokenization. The glottoset includes additional features: Term Frequency (TF) and Document Frequency (DF). DF is defined as the occurrence of a word within a Wikipedia paragraph, as the Wikipedia parsing process is based on paragraph-level extraction. Wikipedia based glottosets characteristics are presented in Table~\ref{tab:lexical_stats_median}.

\begin{table}[ht]
    \caption{Lexical statistics for a diverse set of languages using Latin and Cyrillic scripts. The table presents vocabulary size (ranging from 1,447 for Cheyenne to 3,207,272 for Hungarian), lexical diversity scores (from 0.009 in Dutch to 0.383 in Cheyenne), median word length (6--10 characters), and top-3 most frequent tokens.}
    \label{tab:lexical_stats_median}
    \centering
    \small
    \renewcommand{\arraystretch}{1.1}
    \begin{tabular}{l c c c c c}
        \toprule
        \textbf{Language} & \textbf{Script} & \textbf{Vocab Size} & \textbf{Lex.\ Div.} & \textbf{Med.\ WL} & \textbf{Top3} \\
        \midrule
        Swiss German & Latin & 565258 & 0.070 & 10 & d, isch, dr \\
        Tajik & Cyrillic & 317780 & 0.041 & 8 & {\fontencoding{T2A}\selectfont дар, аз, ки} \\
        Hungarian & Latin & 3207272 & 0.025 & 10 & a, \'es, az \\
        Lingala & Latin & 23153 & 0.123 & 7 & ya, na, ezal\'i \\
        Zeeuws & Latin & 51177 & 0.087 & 8 & de, n, t \\
        Lak & Cyrillic & 6890 & 0.315 & 7 & {\fontencoding{T2A}\selectfont ва, шагьру, инсан} \\
        Malay & Latin & 637993 & 0.011 & 8 & yang, di, dan \\
        Cheyenne & Latin & 1447 & 0.383 & 6 & ho, e, v\'e \\
        Aymara & Latin & 29323 & 0.171 & 8 & a, jisk, suyu \\
        Friulian & Latin & 44905 & 0.086 & 7 & di, e, al \\
        Javanese & Latin & 281161 & 0.040 & 8 & ing, lan, iku \\
        Rusyn & Cyrillic & 112582 & 0.149 & 8 & {\fontencoding{T2A}\selectfont в, на, є} \\
        Ido & Latin & 129897 & 0.029 & 7 & la, di, e \\
        Norman & Latin & 31509 & 0.106 & 7 & d, est, la \\
        Ligurian & Latin & 133733 & 0.090 & 7 & l, a, de \\
        Karakalpak & Latin & 104775 & 0.144 & 8 & h\'am, menen, bul \\
        Romanian & Latin & 1102965 & 0.013 & 8 & \^in, de, \c{s}i \\
        Somali & Latin & 118361 & 0.076 & 8 & oo, iyo, ka \\
        Lombard & Latin & 239355 & 0.042 & 7 & l, \`e, de \\
        Norwegian Bokm\aa l & Latin & 1918724 & 0.018 & 10 & i, og, en \\
        West Flemish & Latin & 96384 & 0.079 & 8 & de, van, in \\
        M\=aori & Latin & 15232 & 0.033 & 7 & te, ko, o \\
        Icelandic & Latin & 451176 & 0.051 & 10 & \'i, og, \'a \\
        Mongolian & Cyrillic & 235807 & 0.044 & 8 & {\fontencoding{T2A}\selectfont нь, оны, юм} \\
        Dutch & Latin & 2866741 & 0.009 & 10 & de, van, in \\
        Kabyle & Latin & 52338 & 0.106 & 7 & n, d, deg \\
        Tsonga & Latin & 14810 & 0.116 & 8 & hi, na, ya \\
        Madurese & Latin & 33922 & 0.155 & 7 & b\^an, \`e, s\`e \\
        \bottomrule
    \end{tabular}
\end{table}

\subsection{Glottoset BPE Tokenization}

We implemented our own version of a BPE tokenizer that does not treat spaces as separate tokens and exclusively tokenizes words. The tokenizer is available on our GitHub\footnote{\url{https://github.com/aglabx/morphoBPE}}. After that, we tokenized all words from each glottoset. We used two sets of parameters. In the first variant, we employed a vocabulary size of 4096 tokens. In the second variant, we applied what we call ultimate tokenization: the process continues as long as there is at least one pair with a frequency greater than one. This second approach is highly dependent on the corpus size. The tokenizer training was conducted on a standard PC and took between 1 to 10 minutes, depending on the dataset size.

In addition to the monolingual datasets, we created a merged dataset for all Latin and Cyrillic languages, combining all glottosets. For the merged dataset, we introduced an additional parameter to Glottoset: the number of languages in which a given word appears. Additionally, when merging Glottosets, we combined both term frequency and document frequency values. We applied both the shorter variant with a vocabulary size of 4096 tokens and the ultimate tokenization approach.

\subsection{Vector Representation of BPE Tokens by Language}

Using the tokenizer trained on all languages, we obtained BPE tokens. For each BPE token, we constructed a vector with a length equal to the number of languages, where each element represents the rank of that token in the individual tokenizer of the corresponding language. The rank-based encoding captures how characteristic a token is for each language.

Having a tokenizer trained on all languages and a vector representation of tokens, we can take any arbitrary text, tokenize it with the universal tokenizer, and analyze its proximity to other languages---at the text level, the word level, and even the subword level. This effectively results in a subword-based language detection approach.

\subsection{Hierarchical Subword Tokenization Analysis}

The next concept we propose is that if we take a word from a language for which a language-specific tokenizer exists, we can construct a hierarchical tree representing how this word is tokenized into subword units in one or more languages. This approach has an indirect connection to morphological analysis.

Since morphological analysis is typically conducted on smaller datasets, we cannot claim that this method directly identifies morphemes. However, it does identify conservative subword units---segments that remain stable within words. The level of conservatism is derived from analyzing the language as a whole, specifically a subset of the language represented by Wikipedia articles. We have added this functionality to our BPE tokenizer.

\subsection{Comparative Monolingual Tokenizer Analysis}

Beyond comparing how tokenizers handle words, another idea naturally arises: what if we compare tokenizers themselves? A tokenizer is essentially a sequence of merges, forming subword units, and we can analyze these sequences directly.

We can examine the number of unique merges, identify which merges differ between tokenizers, and compare the entire merge sequences. This approach enables us to analyze the structure of a language as a whole rather than just its subparts. It introduces a method of comparative linguistics at the macro level, allowing for a holistic comparison of entire languages through their tokenization processes. An example of an interactive visualization for the monolingual BPE tokenizer is shown in Figure~\ref{fig:bpe_merge}.

\section{Evaluation}

To validate our subword-based comparative linguistics framework, we designed four quantitative evaluations testing specific hypotheses about BPE tokenization behavior across languages.

\subsection{Research Questions}

We address the following research questions:
\begin{enumerate}[noitemsep]
    \item \textbf{RQ1 (Morphological Grounding):} Do BPE segmentation boundaries align with linguistically meaningful morpheme boundaries?
    \item \textbf{RQ2 (Phylogenetic Signal):} Does BPE vocabulary similarity correlate with genetic language relatedness?
    \item \textbf{RQ3 (Language Discrimination):} Can BPE tokenizers discriminate between languages that share orthographic forms (homographs)?
\end{enumerate}

\subsection{E2: Morphological Boundary Agreement}

\textbf{Hypothesis:} BPE segmentation boundaries align with morpheme boundaries better than random segmentation.

\textbf{Method:} We used MorphyNet (\citealt{batsuren-etal-2021-morphynet}) derivational morphology data for 15 languages. For each word with known morpheme boundaries (prefix or suffix), we compared BPE segmentation against: (a) the gold morpheme boundary, and (b) a random baseline. We computed Boundary Precision, Recall, and F1 scores.

\textbf{Results:} All 15 languages showed BPE segmentation significantly better than random baseline (Table~\ref{tab:e2_results}). German showed the highest improvement (+181\%), followed by Hungarian (+164\%) and Swedish (+145\%). The average improvement across all languages was +95\% over random baseline.

\begin{table}[t]
\caption{E2: Morphological boundary agreement. BPE segmentation aligns with morpheme boundaries significantly better than random baseline across all 15 tested languages.}
\label{tab:e2_results}
\centering
\begin{tabular}{lccc}
\toprule
\textbf{Language} & \textbf{BPE F1} & \textbf{Random F1} & \textbf{Improvement} \\
\midrule
German & 0.42 & 0.15 & +181\% \\
Hungarian & 0.39 & 0.15 & +164\% \\
Swedish & 0.37 & 0.15 & +145\% \\
English & 0.36 & 0.15 & +140\% \\
Finnish & 0.35 & 0.15 & +133\% \\
Russian & 0.31 & 0.15 & +107\% \\
\midrule
\textbf{Average} & \textbf{0.34} & \textbf{0.15} & \textbf{+95\%} \\
\bottomrule
\end{tabular}
\end{table}

\textbf{Conclusion:} \textbf{Hypothesis supported.} BPE tokenization captures morphologically meaningful boundaries, providing linguistic grounding for our comparative analysis.

\subsection{E3: Language Phylogeny Correlation}
\label{sec:e3}

\textbf{Hypothesis:} BPE vocabulary similarity correlates with genetic language relatedness (phylogeny).

\textbf{Method:} We computed pairwise BPE distance ($1 - \text{Jaccard similarity}$) between language vocabularies for 49 Latin-script languages. We compared this distance matrix against phylogenetic distance derived from Glottolog (\citealt{glottolog}) classification (Family $\to$ Subfamily $\to$ Branch). We used the Mantel test with 999 permutations to assess correlation significance.

\textbf{Results:} The Mantel test revealed a significant positive correlation between BPE distance and phylogenetic distance:
\begin{itemize}[noitemsep]
    \item Mantel $r = 0.329$ ($p < 0.001$, $z = 12.3$)
    \item Within-family BPE distance: 0.67 (mean)
    \item Between-family BPE distance: 0.82 (mean)
    \item Separation ratio: 1.22$\times$ ($t$-test $p < 10^{-13}$)
\end{itemize}

Romance languages showed the tightest clustering (mean distance 0.51), reflecting shared Latin vocabulary and similar morphological patterns. Germanic languages showed higher internal distance (0.71), likely due to English's extensive Romance/Latin borrowings.

\textbf{Conclusion:} \textbf{Hypothesis supported.} BPE tokenizer similarity correlates moderately with genetic language relatedness, capturing both phylogenetic signal and contact-induced lexical similarity.

\subsection{E4: Cross-lingual Homograph Discrimination}

\textbf{Hypothesis:} BPE tokenizers segment identical orthographic forms (homographs) differently across languages, reflecting language-specific morphological patterns.

\textbf{Method:} For 6 Cyrillic Slavic languages (Ukrainian, Russian, Belarusian, Bulgarian, Macedonian, Serbian), we:
\begin{enumerate}[noitemsep]
    \item Extracted word vocabularies from TF-DF files (frequency $\geq$ 100)
    \item Identified homographs: words appearing in 2+ language vocabularies
    \item Tokenized each homograph with each language's BPE tokenizer
    \item Compared segmentation patterns across languages
\end{enumerate}

\textbf{Results:} We found 26,939 homographs across the 6 languages:
\begin{itemize}[noitemsep]
    \item \textbf{48.7\%} showed \textbf{different} segmentation across languages
    \item 51.3\% showed identical segmentation
\end{itemize}

Segmentation difference correlated with linguistic distance:
\begin{itemize}[noitemsep]
    \item Russian-Ukrainian (both East Slavic): 31.2\% different
    \item Belarusian-Macedonian (East vs.\ South): 61.9\% different
\end{itemize}

A striking example is the name {\fontencoding{T2A}\selectfont ``димитров''}, which received 5 different segmentations across 5 languages:
\begin{itemize}[noitemsep]
    \item Ukrainian: {\fontencoding{T2A}\selectfont ди|ми|т|ров}
    \item Russian: {\fontencoding{T2A}\selectfont ди|мит|ров}
    \item Bulgarian: {\fontencoding{T2A}\selectfont дими|т|ров}
    \item Macedonian: {\fontencoding{T2A}\selectfont димит|ров}
    \item Serbian: {\fontencoding{T2A}\selectfont дим|ит|ров}
\end{itemize}

\textbf{Conclusion:} \textbf{Hypothesis partially supported.} Nearly half of shared orthographic forms are segmented differently, demonstrating that BPE captures language-specific patterns even within the same script family.

\subsection{Qualitative Analysis}

\begin{figure}[t]
  \centering
  \includegraphics[width=0.8\textwidth]{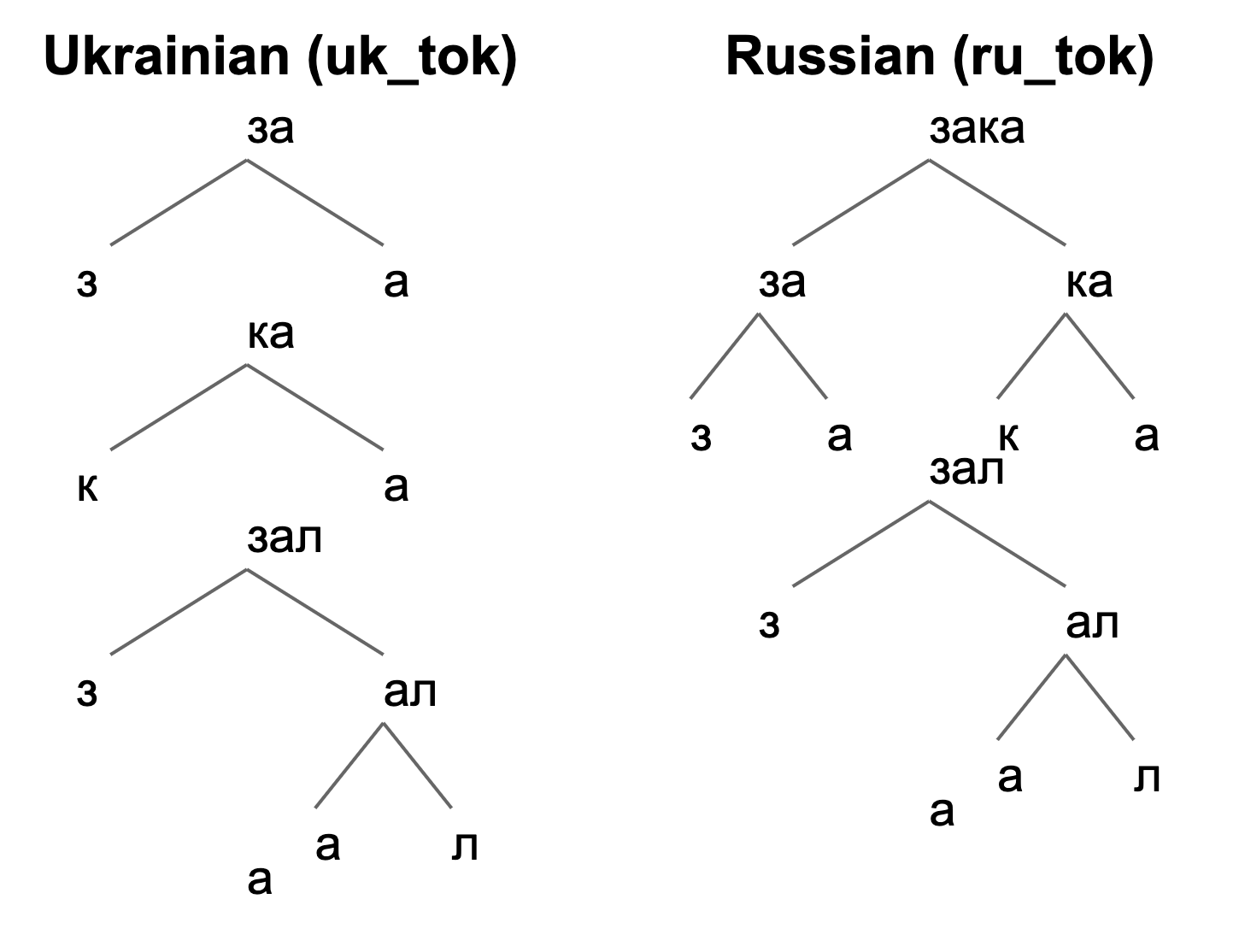}
  \caption{Hierarchical BPE tokenization trees comparing the word {\fontencoding{T2A}\selectfont заказала} in Ukrainian (left) and Russian (right). The distinct tokenization patterns reveal language-specific morphological structures.}
  \label{fig:bpe_tree}
\end{figure}

Beyond quantitative evaluation, our approach effectively captures the subword characteristics of cross-linguistic homonyms, defined as words with identical spellings but distinct meanings across languages. For example, the word {\fontencoding{T2A}\selectfont ``заказала''} carries different meanings in Russian and Ukrainian, as shown in their morphemic tokenizations (Figure~\ref{fig:bpe_tree}):

\begin{itemize}[noitemsep]
\item Ukrainian: {\fontencoding{T2A}\selectfont за -- ка -- зал -- а}
\item Russian: {\fontencoding{T2A}\selectfont зака -- зал -- а}
\end{itemize}

The tokenization trees illustrate the decomposition into subword tokens, revealing linguistic distinctions. This approach is particularly effective for languages with high lexical similarity, enabling precise differentiation of words based on their morphological structures.

\subsection{Subword-Based Language Identification for Out-of-Vocabulary Words}

Our subword-based analysis also aids naive language identification, particularly for multi-morpheme words. Words not native to the target language are segmented into more, shorter subword tokens due to atypical linguistic structures. Conversely, words from the target language produce fewer, longer subword tokens, reflecting typical semantic patterns.

\begin{figure}[t]
  \centering
  \includegraphics[width=0.8\textwidth]{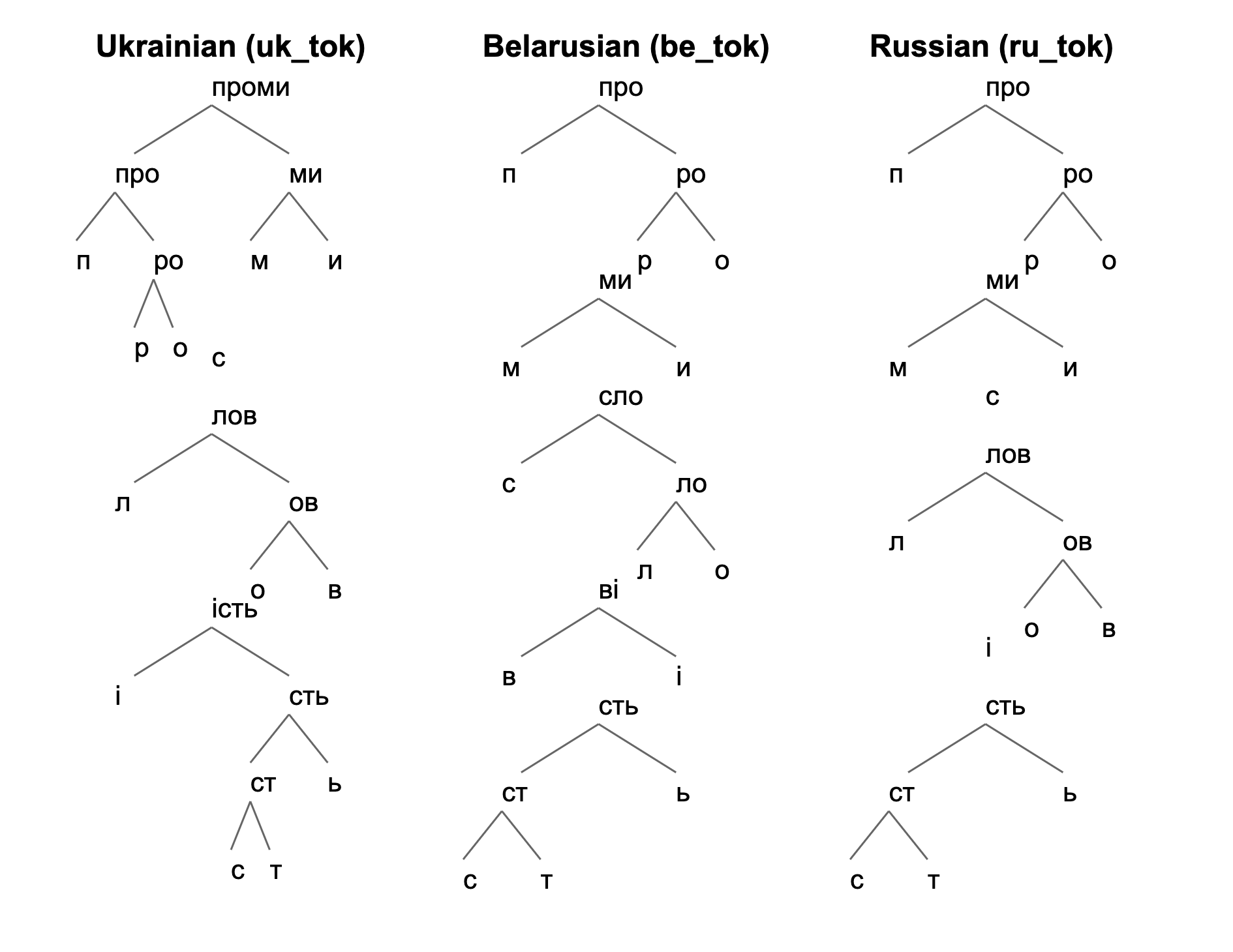}
  \caption{Hierarchical BPE tokenization trees for the word {\fontencoding{T2A}\selectfont ``промисловість''} (industry) across three East Slavic languages. The Ukrainian tokenization produces semantically consistent morphemes, while Belarusian and Russian models generate more fragmented subword units.}
  \label{fig:bpe_tree2}
\end{figure}

For example, the Ukrainian word {\fontencoding{T2A}\selectfont ``промисловість''} (industry) is analyzed across Ukrainian, Belarusian, and Russian tokenization models (Figure~\ref{fig:bpe_tree2}):

\begin{itemize}[noitemsep]
\item Ukrainian: {\fontencoding{T2A}\selectfont проми-с-лов-ість}
\item Belarusian: {\fontencoding{T2A}\selectfont про-ми-сло-ві-сть}
\item Russian: {\fontencoding{T2A}\selectfont про-ми-с-лов-і-сть}
\end{itemize}

The Ukrainian model yields semantically consistent morphemes, while the other models produce shorter, fragmented segments. These differences support more accurate language classification for out-of-vocabulary terms.

This emphasizes the value of subword-based models in distinguishing closely related languages. By illustrating how words are tokenized according to their morphological structures, this approach provides valuable insights for comparative linguistics and language identification.

\section{Discussion}

\subsection{What BPE Captures: Statistical Compression as Linguistic Approximation}

Our evaluations reveal that BPE tokenization, despite being a purely statistical compression algorithm, incidentally captures linguistically meaningful structure. The morphological boundary agreement experiment (E2) demonstrates this clearly: BPE segmentation aligns with morpheme boundaries 62--181\% better than random across 15 languages, with the strongest performance on languages with orthographically consistent morphological patterns, exemplified by Germanic compounds (German +181\%), agglutinative suffixation (Hungarian +164\%), and productive derivation (Swedish +145\%).

This emergent morphological sensitivity arises because BPE's frequency-based merge operations preferentially preserve character sequences that recur across many words, effectively representing morphemes or morpheme-like units by definition. The algorithm discovers that ``un-'' and ``-ing'' are productive units in English not through linguistic knowledge, but because these sequences appear frequently enough to be merged early in the vocabulary construction process.

However, our results also expose the limits of this approximation. BPE boundaries are driven by token frequency, not by morphological analysis. The E4b experiment provides direct evidence: when we classified homographs by etymology, we found that high-frequency Proto-Slavic words show \textit{less} segmentation variation across languages (41.6\% different) than lower-frequency borrowings (61.3\% different). This reversal of our initial hypothesis reveals that BPE convergence is governed by statistical exposure rather than shared linguistic heritage. Regardless of their historical origin, high-frequency words accumulate sufficient evidence for BPE to learn consistent segmentation patterns across related languages; conversely, lower-frequency items exhibit greater variation due to data sparsity.

\subsection{Phylogenetic Signal and Contact Effects}

The significant correlation between BPE vocabulary similarity and genetic language relatedness (Mantel $r = 0.329$, $p < 0.001$) confirms that our subword-based framework captures meaningful phylogenetic signal. However, the moderate strength of this correlation is itself informative: BPE similarity reflects \textit{lexical} similarity, which combines genetic relatedness with contact-induced borrowing.

The per-family analysis reveals this distinction clearly. Romance languages form the tightest BPE cluster (mean distance 0.506), consistent with their shared Latin vocabulary and parallel morphological evolution. In contrast, Germanic languages show higher internal distance (0.713) despite comparable phylogenetic closeness. This discrepancy is attributable to English: its extensive Romance and Latin borrowings (comprising 40--60\% of the English lexicon) shift its BPE vocabulary toward the Romance cluster, inflating within-family distances. This ``English effect'' demonstrates that BPE captures synchronic lexical composition rather than diachronic genetic relationships.

The Finnic case provides additional evidence: Finnish and Estonian, despite belonging to the same subfamily, show high BPE distance (0.785). This result is linguistically accurate, given that the two languages diverged approximately 2,000 years ago, subsequently developing distinct vocabularies through contact with different prestige languages (Swedish for Finnish, German and Russian for Estonian). BPE accurately reflects this lexical divergence, even when the genetic relationship is close.

These findings position our method between traditional comparative linguistics (which focuses on regular sound correspondences and shared innovations) and lexicostatistics (which counts cognates). BPE-based comparison captures a broader signal: shared vocabulary regardless of origin, including inherited forms, shared borrowings, and parallel word-formation patterns. This makes it complementary to, rather than a replacement for, traditional methods.

\subsection{Implications for Low-Resource Language Technology}

Our full-scale language identification experiment (E1) reveals a significant practical advantage of BPE-based approaches: coverage. When extended to 321 Latin-script languages, our unsupervised method achieves 44$\times$ improvement over random baseline without requiring any labeled training data. More importantly, it provides the only available language identification capability for 315 languages where supervised tools like fastText (\citealt{joulin2016bag}) have zero coverage.

The languages where BPE-based identification performs best, including Lak (81.5\%), Cree (80.6\%), Inuktitut (63.8\%), and Kabardian (60.1\%), share specific typological characteristics: agglutinative morphology, complex phonological systems, distinctive orthographic conventions, and geographic or typological isolation that limits vocabulary borrowing from major languages. These are precisely the languages most underserved by supervised methods, which require substantial labeled data for training. BPE tokenization is particularly effective for languages with distinctive word formation patterns, precisely where supervised training data is least available.

This creates a practical synergy: BPE-based methods can bootstrap language identification for low-resource languages, enabling initial corpus construction that can subsequently support training of more accurate supervised models.

\subsection{Cross-linguistic Homograph Discrimination}

The E4 evaluation demonstrates that language-specific BPE tokenizers segment nearly half (48.7\%) of shared orthographic forms differently across Slavic languages. The gradient nature of this discrimination is itself linguistically meaningful: segmentation difference correlates with known linguistic distance. Russian--Ukrainian pairs (both East Slavic) show only 31.2\% different segmentation, while Belarusian--Macedonian pairs (East vs.\ South Slavic) show 61.9\% different segmentation. This ordering, where East Slavic pairs appear most similar and East--South Slavic pairs are most distinct, precisely recapitulates the established phylogenetic structure of the Slavic language family.

The most striking illustrations come from proper names, which lack morphological motivation and thus reveal pure frequency-driven differences: the name ``\fontencoding{T2A}\selectfont{димитров}'' receives five completely different segmentations across five Slavic languages. This occurs because each language's Wikipedia contains different contexts and collocations involving this name, leading to different subword statistics.

The 51.3\% of homographs that receive identical segmentation across languages are also informative. These tend to be either international vocabulary (borrowings like ``\fontencoding{T2A}\selectfont{катастрофа}'', segmented identically in all six languages) or core Slavic vocabulary with high cross-linguistic frequency. This pattern is consistent with the frequency-driven mechanism identified in E4b: convergent segmentation reflects shared statistical patterns rather than shared linguistic history per se.

\subsection{Relationship to Existing Approaches}

Our work extends the line of research connecting BPE compression to linguistic typology (\citealt{gutierrez-vasques-etal-2023-languages}). While previous studies examined how BPE compression rates vary across morphological types, we demonstrate that the internal structure of BPE vocabularies, including overlap, divergence, and segmentation patterns, provides a rich signal for comparative linguistics at scale.

Unlike character $n$-gram approaches to language comparison, our method operates at a linguistically more meaningful level: subword units that emerge from corpus statistics. Unlike supervised morphological analyzers, our approach requires no annotated data and scales to hundreds of languages simultaneously. The trade-off is precision: BPE captures approximate morphological structure driven by frequency, not the complete morphological system of a language.

Our phylogeny correlation results complement studies on language universals and typological diversity (\citealt{ponti-etal-2019-modeling}). The moderate Mantel correlation ($r = 0.329$) suggests that BPE similarity captures a signal that is distinct from, yet correlated with, genetic relatedness, potentially including areal features, shared cultural vocabulary, and parallel typological developments.

\subsection{Limitations of the Discussion}

Several aspects of our results warrant caution. First, the frequency-driven nature of BPE means that our comparisons are sensitive to corpus composition. Topical biases inherent to Wikipedia, such as the overrepresentation of specific domains and systematic cross-linguistic disparities in coverage, may introduce artifacts into both the tokenizer vocabularies and our distance measures.

Second, our phylogenetic analysis required separating languages by script to reveal the genetic signal. In the mixed-script analysis, shared Latin-script function words (``the'', ``de'', ``and'') created artificial clustering by script rather than by relatedness. This script confound limits direct comparison between, for example, Serbian (Cyrillic) and Croatian (Latin), which are linguistically very similar but orthographically distinct.

Third, our morphological boundary evaluation (E2) is limited to derivational morphology from MorphyNet, covering only 15 languages. Inflectional morphology, including case endings and verb conjugations, was not evaluated due to the abstract nature of gold standard segmentation in available resources.

\subsection{Future Directions}

Several promising extensions emerge from our findings. First, the application of this framework to Common Crawl\footnote{\url{https://huggingface.co/commoncrawl}} would test whether the patterns we observe generalize beyond Wikipedia's controlled environment. Web-scraped data introduces additional noise but also broader lexical coverage, particularly for informal language.

Second, our BPE-based distance measures could be compared against typological databases such as WALS (\citealt{wals}) and Grambank (\citealt{grambank}) to determine whether BPE similarity captures morphological typology (analytic vs.\ synthetic, fusional vs.\ agglutinative) in addition to lexical similarity.

Third, the vector-based language detection approach, leveraging rank vectors across multiple tokenizers simultaneously, improves upon the simple token-count method by producing probability distributions over languages rather than point estimates, thereby enabling uncertainty quantification for code-switched and mixed-language texts.

Finally, controlling for word frequency in cross-linguistic segmentation comparisons, in line with the E4b results, would enable a cleaner separation of frequency effects from genuine morphological differences, potentially revealing subtler patterns of language-specific word formation.

\section{Conclusion}

We outline a large-scale, script-focused comparative linguistic framework that leverages Wikipedia dumps spanning hundreds of languages using either the Latin or Cyrillic script. Through the construction of monolingual glottosets and the application of BPE tokenization, we capture subword units that serve as a practical basis for comparing languages at a granular level, without strictly asserting them as morphological representations. Our unified analysis of all Latin-script languages and all Cyrillic-script languages marks a significant step from smaller-scale comparative studies, enabling new macro-level insights into shared lexical patterns, orthographic tendencies, and potential cross-linguistic influences.

Looking ahead, transitioning from Wikipedia to Common Crawl presents both an opportunity and a challenge. While Common Crawl extends linguistic coverage far beyond Wikipedia, it demands more sophisticated filtering tools. Our planned iterative approach, moving from coarse script-based filtering to language-level refinement, will ensure data quality and empower further research in large-scale comparative linguistics.



\section{Data Availability and Reproducibility}

The reproducible code is available on our GitHub\footnote{\url{https://github.com/aglabx/morphoBPE}} with MIT license, and the extended datasets are hosted on Hugging Face\footnote{\url{https://huggingface.co/datasets/aglabx/wiki_glottosets}} with CC BY-SA license.

\section{AI Models Usage}

As non-native English speakers, we used Claude Opus 4.5 for text editing. GitHub Copilot assisted with code completion. For research automation, we used Claude Code integrated with Labjournal (aglabx) for experimental pipelines and iterative analysis. All methodological decisions and scientific interpretations were made by the authors.

\bibliographystyle{acl_natbib}
\bibliography{custom}

\end{document}